\documentclass{article}

\usepackage{titlesec}
\usepackage[margin=1.3in,includefoot]{geometry}
\usepackage[applemac]{inputenc}
\usepackage{graphicx}
\usepackage{amsmath}
\usepackage{algorithm}
\usepackage{algorithmic}
\usepackage{float}

\titleformat{\section}[hang]{\large\scshape}{\thesection.}{1em}{}

\begin{document}
	\begin{flushleft}
	\textsc{\LARGE Vision-based Robotic Arm Imitation by Human Gesture}
	\end{flushleft}
	{\normalsize\bfseries Cheng Xuan, Zhiqiang Tang, Jinxin Xu}\\
	{Tsinghua University, University of Alberta}\\
	{Beijing, People's Republic of China, Edmonton, Canada}\\
	\texttt{xc17@mails.tsinghua.edu.cn, ztang4@ualberta.ca, jinxin3@ualberta.ca}
	\footnote{Authors contributed equally}\\[12pt]

	\begin{center}
	\textsc{\large Abstract}
	\end{center}

	\begin{changemargin}{0.5in}{0.5in}
	One of the most efficient ways for a learning-based robotic arm to learn to process complex tasks as human, is to directly learn from observing how human complete those tasks, and then imitate. Our idea is based on success of Deep Q-Learning (DQN) algorithm according to reinforcement learning, and then extend to Deep Deterministic Policy Gradient (DDPG) algorithm. We developed a learning-based method, combining modified DDPG and visual imitation network. Our approach acquires frames only from a monocular camera, and no need to either construct a 3D environment or generate actual points. The result we expected during training, was that robot would be able to move as almost the same as how human hands did. \\[6pt]
	\end{changemargin}

\section{Introduction}\label{sec:intro}
Designing a robotic arm control system which aims at everyday tasks is one of the biggest challenges among current robot researches.\\[7pt]
Recently, as Machine Learning being more and more popular, Deep Reinforcement Learning (DRL) have performed decently well dealing with some general and simple tasks, such as letting robotic arms learn to open doors with different kinds of handles after a lot of training (Yahya et all., 2016). However, the idea of letting robots work in vary situations and try to solve a same task in all those different cases is sort of intuitive. The main problem is that such idea might require several months' training (Levine et all., 2016), and therefore it may not as practical as we expected in everyday tasks.\\[7pt]
Our primary goal is to let the robotic arm itself learn from human, and to touch, push or even pick up items according to how exactly human do. This approach is able to significantly decrease the number of training steps by visually imitate human demonstrations. Further more, unlike other approaches of robot controlling with DDPG, which applied a random exploration noise while selecting an action, our algorithm allows robot to explore on purpose according to human movement, so that it will be more efficient in finding the best solution. A primary contributions we have done is that we modified DDPG algorithm, where we applied a Heuristic value to help to choose next actions, and we call that a {\em Heuristic DDPG}. This approach significantly reduce those unnecessary explorations accelerate our training efficiency, and since this approach is able to help robot to learn from human, our robot can perform in a human-like way, which makes it practical dealing with everyday tasks.\\

\section{Algorithm}
Our method has two components. To generate heuristic values, we built a visual imitation neural network which used a deep convolutional neural network (CNN) to compare the similarity which is represented by our heuristic values, between human movement and robotic move trajectories. Moreover, we used a modified DDPG (Lillicarp et all., 2016) to continuously control the robot based on current frame from the camera and heuristic values acquired from imitation neural network.\\[7pt]
\begin{figure}[H]
	\center
 	\includegraphics[width=\linewidth]{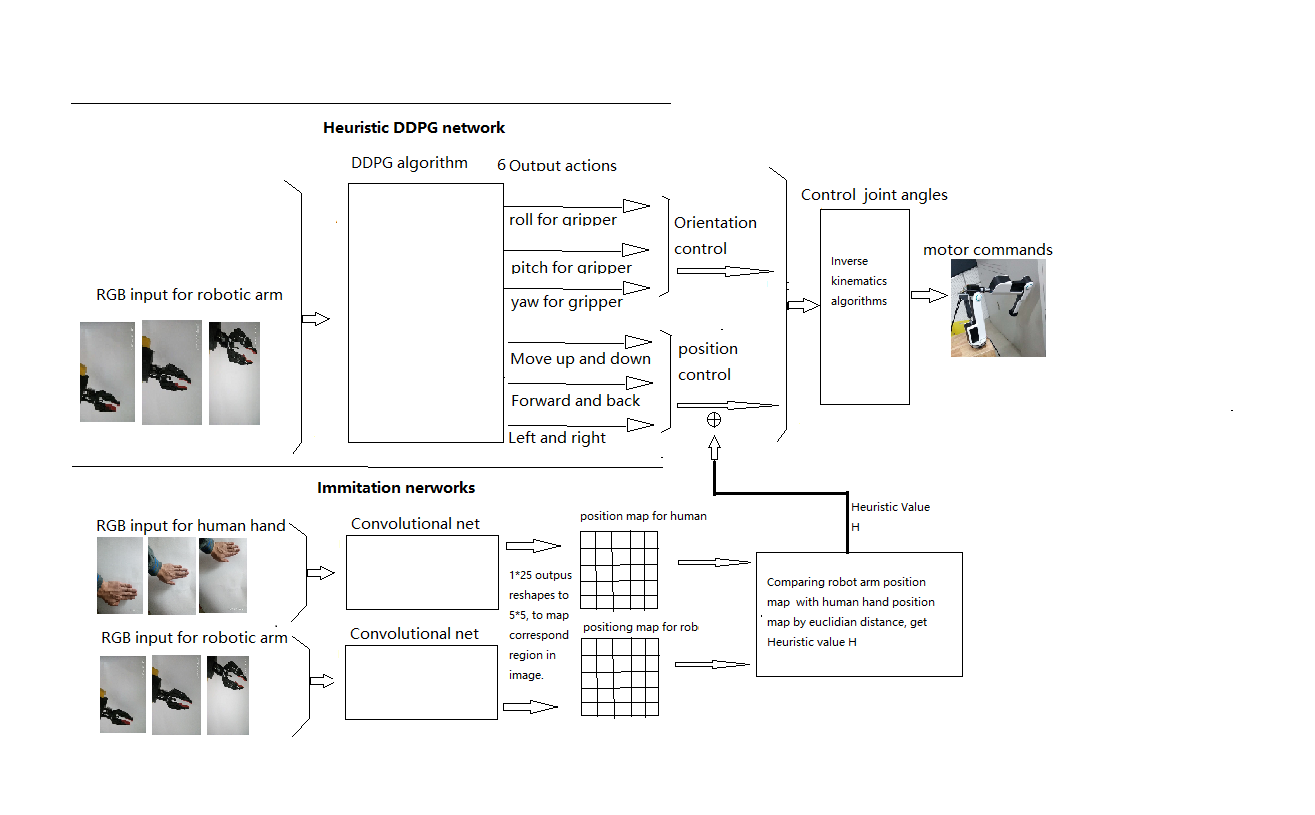}
  	\caption{Overview Framework including Heuristic DDPG network and imitation neural network}
\end{figure}
To make the overall framework more clear, let's go through this in following procedures. we firstly moved our hands from up to down and from left to right in front of the camera and recorded a video. Then, we moved our robot arm and let it process what human just did. Next, we built a convolutional neural network which took both robot and human hands' frames of motion as input and locations of either the robotic arm or human hands in frames as output. After this, we were able to start to train the network and acquire heuristic values. After training imitation network, we can start testing our model on some simple tasks-pushing a button in front of the robot camera for example. Each frame gained from the camera will be feed into the imitation network and get estimated hands positions, and then we push every hand position into a stack which will turn into a sequence of data and form a trajectory. Finally, the robotic arm will start to move under the observation of camera and each frame will go through two parts. One is the imitation network to estimate current position of robot and then get a trajectory, the other is DDPG network to control robot based on its current state. Before actuating the robot, our system will compare the similarity of human hands and robot trajectories, and then generate a value of difference which is called heuristic action value. The actual movement for robot will be sum of heuristic value and DDPG action output, which actually will be able to reduce unnecessary exploration.\\

\subsection{\itshape\normalsize Data Acquisition and Preprocessing}
Since our goal is to imitate human hands movement using our robotic arm, we firstly need to do a space mapping with robot and human. Unlike building a transformation matrix to describe the relative position of human hands and robot, we used the recorded images instead to acquire those positions which are represented by image coordinates.\\[7pt]
\begin{figure}[H]
	\center
 	\includegraphics[width = 12cm, height = 9cm]{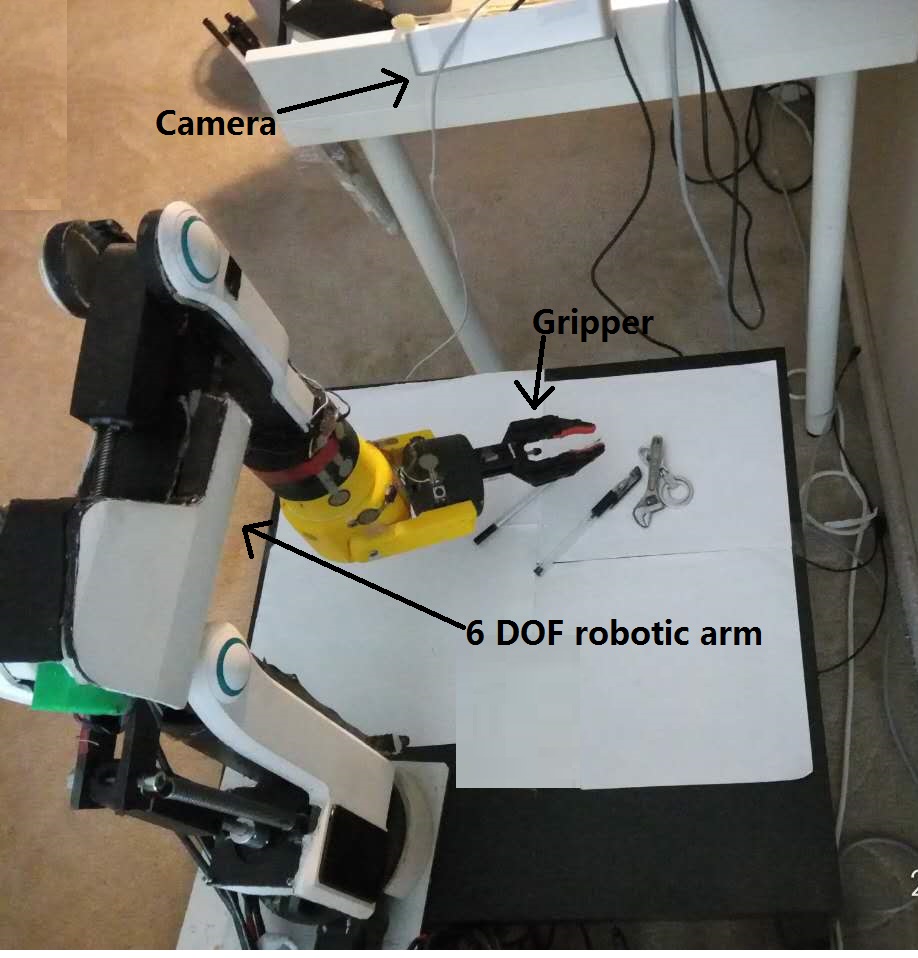}
  	\caption{Environment of data acquisition}
\end{figure}
We manually Tagged those images, and then tagged names by coordinate values of hands or robot. Finally we fed these images to the imitation net, and start to train our network. To increase the number of training datasets and robustness of the neural net, we did some image preprocessing-we increased the brightness and decreased the contrast, normalized standard deviation, and added salt and pepper noise on original images. Image translation, rotation, and flipping were not performed since the position information is important in this case. \\

\begin{figure}[H]
	\center
 	\includegraphics[width = 12cm, height = 9cm]{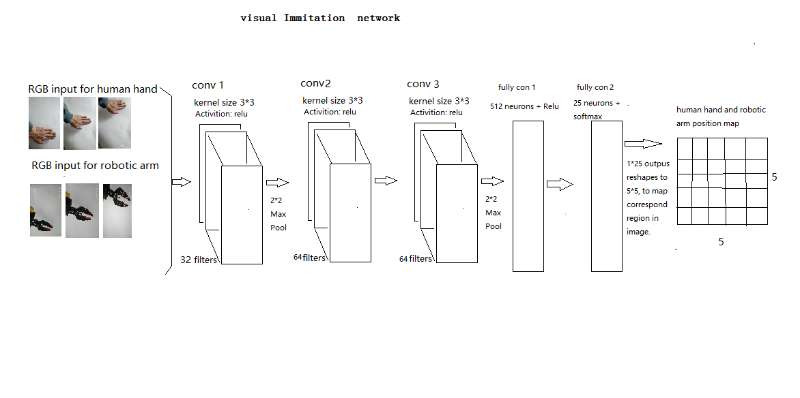}
  	\caption{Imitation Network}
\end{figure}

\subsection{\itshape\normalsize Imitation Network}
Our visual imitation network is a CNN based network, which has three convolution layers, two max pooling layers, and two fully connected layers. CNNs are commonly used in image classifications, and in our approach, it is used for position estimation of our robot and hands. The output is a type of one-hot encoding and has 25 values, we reshaped the 25*1 one dimensional output array into 5*5 two dimensional matrix to stand for 25 region in the image. \\
\begin{figure}[H]
	\center
 	\includegraphics[width = 3cm, height = 3cm]{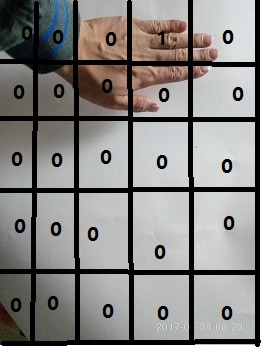}
  	\caption{The black grid stands for output matrix according to imitation network, since output is one-hot encoding the index of 1 in the matrix corresponding positions of hands or robot gripper in images. Therefore the estimated coordinate for the hand in image shown is (0,3)
}
\end{figure}

\subsection{\itshape\normalsize Heuristic DDPG net}
DDPG algorithm is successfully used in continuously-especially high dimensional which in our case is $3^6$ different set of actions-control applications. Inspired from underlying Deep Q-Learning to the continuous action spaces. DDPG is an actor-critic(Sutton et all., 2015), model-free algorithm based on the deterministic policy gradient (Lillicarp et all., 2016). Since DDPG needs to do random exploration to find optimal actions, and however, continuous actions usually have large action spaces. As a result, training a DDPG network usually needs millions of steps, which may not be a problem for simulated tasks such as playing atari games, but will be problematic in real world application. For example, controlling a robotic arm which has seven or eight rollers. To conquer this challenge, we came up with an idea called {\em Heuristic exploration}.\\

\begin{algorithm}[H]
\caption{Imitation network and modified DDPG}
\begin{algorithmic}[1]
\STATE $\text{Initialize imitation network }H_{net}(\theta) \text{ }Map_{hand}$
\STATE $\text{Load imitation training images }image_{robot}\text{ and }image_{hand}$
\FOR{$\text{i from 1 to maximum number of steps}$}
\STATE $\text{Train imitation network }H_{net}(image_{robot}, image_{hand}, \theta)$
\ENDFOR
\STATE $Imageset_{human}\text{ consists of series of images recorded by camera}$
\FOR{$\text{i in range of size(}Imageset_{human}\text{)}$}
\STATE $Map_{hand}\text{[i] = }H_{net}(imageset_{human}\text{[i]},\theta)$
\ENDFOR
\STATE $\text{Randomly initialize critic network }Q(s,a|\theta^Q)\text{ and actor } \mu(s|\theta^\mu)\text{ with weight }\theta^Q\text{ and }\theta^\mu$
\STATE $\text{Initialize target network }Q^\prime\text{ and }\mu^\prime\text{ with weights }\theta^{Q^\prime}\gets\theta^Q,\theta^{\mu^\prime}\gets\theta^\mu$
\STATE $\text{Initialize replay buffer }R$
\FOR{$\text{episode = 1,}M$}
\STATE $\text{Initialize a random process {\itshape N} for action exploration}$
\STATE $\text{Receive initial observation state }s_1$
\FOR{$\text{t = 1,}T$}
\STATE $image_{robot}\gets\text{from camera}$
\STATE $Map_{robot}\text{ = }H_{net}(image_{robot}, \theta)$
\STATE $H_{val}\text{ = Euclidian(}Map_{robot}\text{,}Map_{hand}\text{[t])}$
\STATE $\text{Select action }a_t\text{ = }\mu(s_t,\theta^\mu)\text{ + }H_{val}$
\STATE $\text{Execute action }a_t\text{ and observe reward }r_t\text{ new state }s_{t+1}$
\STATE $\text{Store transition (}s_t,a_t,r_t,s_{t+1}\text{) in }R$
\STATE $\text{Sample a random minibatch of {\itshape N} transitions (}s_i,a_i,r_i,s_{i+1}\text{) from }R$
\STATE $\text{Set }y_i\text{ = }r_i\text{ + }\gamma Q^\prime(s_{i+1},\mu^\prime(s_{i+1}|\theta^{\mu^\prime})|\theta^{Q^\prime})$
\STATE $\text{Update critic by minimizing the loss: }L = \frac{1}{N}\sum_i(y_i - Q(s_i, a_i|\theta^Q))^2$
\STATE $\text{Update the actor policy using the sampled policy gradient: }$
\begin{equation*} 
\nabla_{\theta^\mu}J \approx \frac{1}{N}\sum_i\nabla_aQ(s,a|\theta^Q)|_{s=s_i,a=\mu(s_i)}\nabla_{\theta^\mu}\mu(s|\theta^\mu)|_{s_i}
\end{equation*}
\STATE $\text{Update the target networks: }\theta^{Q^\prime}\gets\tau\theta^Q+(1-\tau)\theta^{Q^\prime}$
\STATE $\text{Update the target networks: }\theta^{\mu^\prime}\gets\tau\theta^\mu+(1-\tau)\theta^{\mu^\prime}$
\ENDFOR
\ENDFOR
\end{algorithmic}
\end{algorithm}

A major problem of learning in continuous action spaces using reinforcement learning is useless exploration. The standard DDPG algorithm uses an exploration policy which adds some random value of noise while selecting actions,
\begin{equation}
\mu^\prime(s_t) = \mu(s_t|\theta^\mu) + N
\end{equation}
Where $\mu^\prime$ is exploration policy network and N is the noise value. Theoretically, DDPG is a random search algorithm and therefore unexceptionally inefficient. To improve the efficiency while exploring, we changed selection of actions based on,
\begin{equation}
\mu^\prime(s_t) = \mu(s_t|\theta^\mu)  + H_t
\end{equation}
where $H_t$ is the heuristic value which is the result of comparing human and robot trajectories passing through our imitation network.

\section{\large References}
Yahya, Ali, Li, Adrian, Kalakrishnan, Mrinal, Chebotar, Yevgen and Levine Sergey. Collective Robot Reinforcement Learning with Distributed Asynchronous Guided Policy Search. {\em arXiv:1610.00673.}
\\[7pt]
Levine, Sergey, Pastor, Peter, Krizhevsky, Alex, Quillen, Deirdre. Learning Hand-Eye Coordination for Robotic Grasping with Deep Learning and Large-Scale Data Collection. {\em arXiv:1603.02199}
\\[7pt]
P. Lillicrap, Timothy, J. Hunt, Jonathan, Pritzel, Alexander et all. Continuous Control with Deep Reinforcement Learning. {\em arXiv:1509.02971v5}
\\[7pt]
S.Sutton, Richard, G.Barto, Andrew. Reinforcement Learning: An Introduction second edition. In {\em Policy Approximation}, p227-p230, 2015
\\[7pt]

\end{document}